\documentclass[twocolumn,english]{revtex4-1}
\pdfoutput=1
\usepackage[T1]{fontenc}
\usepackage[utf8]{inputenc}
\usepackage{geometry}
\usepackage{amsmath}
\usepackage{graphicx}

\makeatletter
\@ifundefined{showcaptionsetup}{}{%
 \PassOptionsToPackage{caption=false}{subfig}}
\usepackage{subfig}
\makeatother

\usepackage{babel}
\begin{document}

\title{Derivative Based Focal Plane Array Nonuniformity Correction}

\author{G. Ness, A. Oved, I. Kakon}

\affiliation{Electro Optics Department, RAFAEL PO Box 2250 (39) Haifa 31021, ISRAEL}
\begin{abstract}
This paper presents a fast and robust method for fixed pattern noise
nonuniformity correction of infrared focal plane arrays. The proposed
method requires neither shutter nor elaborate calibrations and therefore
enables a real time correction with no interruptions. Based on derivative
estimation of the fixed pattern noise from pixel sized translations
of the focal plane array, the proposed method has the advantages of
being invariant to the noise magnitude and robust to unknown camera
and inter-scene movements while being virtually transparent to the
end-user.
\end{abstract}


\maketitle
Infrared focal-plane arrays (FPA) are used in a plethora of imaging
systems designed for various applications. Although FPA fabrication
techniques are constantly improving, their performance is still greatly
affected by fixed pattern noise (FPN). This is in principle an undesired
spatially varying bias and gain terms (higher order terms are usually
negligible). While each detector’s response (gain term) is usually
temporally constant, its bias term tends to drift significantly over
time due to temperature variations of the FPA and its surroundings.
FPN in infrared imaging systems severely deteriorates image quality
which affects both human observer and machine vision based tasks,
and therefore must be addressed accordingly. Over the years, many
algorithms and calibration schemes were introduced in order to estimate
and correct FPN. They can be categorized into two groups: calibration
based and scene based algorithms.

Calibration based algorithms usually provide satisfactory results,
they require either long and costly calibration or repetitive interruption
to the imaging process. The most widely used calibration based technique
\cite{key-1} which known as one point correction, is achieved by
placing a uniformly distributed radiation source (opaque shutter,
de-focusing etc.) in front of the FPA, and by doing so obscuring the
scene from the FPA. Since the radiation is uniformly distributed across
the FPA, any residual pattern at this stage is attributed to the unknown
bias and is easily compensated by subtracting it from all following
frames.

More accurate correction can be achieved by compensating for the gain
nonuniformity in addition to the bias nonuniformity. In contrast to
the temporally varying bias term, the gain term is usually constant
and can be calibrated once. More sophisticated calibration schemes
\cite{key-3,key-5,key-8,key-10,key-12,key-17,spie95 on least square fit,frequency domain }
are also available which enable to maintain low FPN levels for relatively
long operation times. The main drawback of those schemes is the long
and costly calibration process involved.

A second group of FPN estimation techniques is the scene based nonuniformity
correction (NUC) \cite{kumar algorithms,chinese sbn,key-1-1}. In
principle, scene based NUC methods rely on a sequence of frames taken
at different imaging conditions (scene change, varying integration
time, varying the imaging direction etc.). Since the FPN and the imaged
scene are uncorrelated, it is possible to algorithmically separate
the scene from the FPN. Naturally, the scene based approach is favorable
since it neither requires long calibrations nor interfere with the
continuous imaging of the scene. \cite{key-18,kalman filter} show
that the mean and the standard deviation of the signal calculated
over a large collection of frames at each detector are its offset
and gain respectively. In \cite{spie13 constant statistics}, the
advantage of constant statistics is taken for estimating the FPN.
In \cite{key-20}, a neural network approach and retina like processing
techniques suggested to estimate the FPN. Different approach \cite{key-26,key-27}
uses frames produced by dithering the detector in a known pattern.
Our approach can be classified as a scene based paper approach, yet
it differs from a systemic standpoint, as it enables accurate and
robust reconstruction of the FPN from its derivatives. The FPN derivatives
are estimated using frames captured during FPA translations, or during
angular movements of the entire imaging system.

In this paper, we describe the model of FPN reconstruction review
numerical simulation and present experimental results of the algorithm. 

Let $\widetilde{R}_{n}\left(i,j\right)$ be the raw value of the pixel
$\left[i,j\right]$ located at the $i^{th}$ row and $j^{th}$ column
of the $n^{th}$ frame. A simple imaging model for pixel $\left[i,j\right]$
can be
\begin{equation}
\widetilde{R}_{n}\left(i,j\right)=\phi_{n}\left(i,j\right)\cdot g_{n}\left(i,j\right)+\widetilde{o}_{n}\left(i,j\right)+\eta\;,\label{eq:imaging model}
\end{equation}
where $\phi_{n}\left(i,j\right)$ is the radiance emitted from the
scene, integrated over the pixel’s active area within the frame integration
time. $g\left(i,j\right)$ and $\widetilde{o}_{n}\left(i,j\right)$
describe the pixel gain and offset respectively. $\eta$ is the temporal
noise term, which will be neglected in further calculations for simplicity
of description and will be reduced by temporal averaging (the distortion
caused by temporal noise was evaluated by simulation). We assume that
the FPN estimation process is short enough so that the offset term
is considered temporally constant during the correction process, that
is $\widetilde{o}_{n}\left(i,j\right)\equiv\widetilde{o}\left(i,j\right)$.
We also assume that the gain term is known from previous calibration
stage, therefore we can simplify Eq. \ref{eq:imaging model} by compensating
for the gain nonuniformity. This is achieved by multiplying both sides
of Eq. \ref{eq:imaging model} by $g\left(i,j\right)^{-1}$ :

\begin{equation}
R_{n}\left(i,j\right)=\phi_{n}\left(i,j\right)+o\left(i,j\right)\;,\label{eq:gain compensated model}
\end{equation}
where 
\[
\begin{cases}
R_{n}\left(i,j\right)\equiv\widetilde{R}_{n}\left(i,j\right)\cdot g\left(i,j\right)^{-1} & \,\\
o\left(i,j\right)\equiv\widetilde{o}\left(i,j\right)\cdot g\left(i,j\right)^{-1} & \,
\end{cases}\;.
\]

\begin{figure}[!tb]
\begin{centering}
\includegraphics[width=0.75\columnwidth]{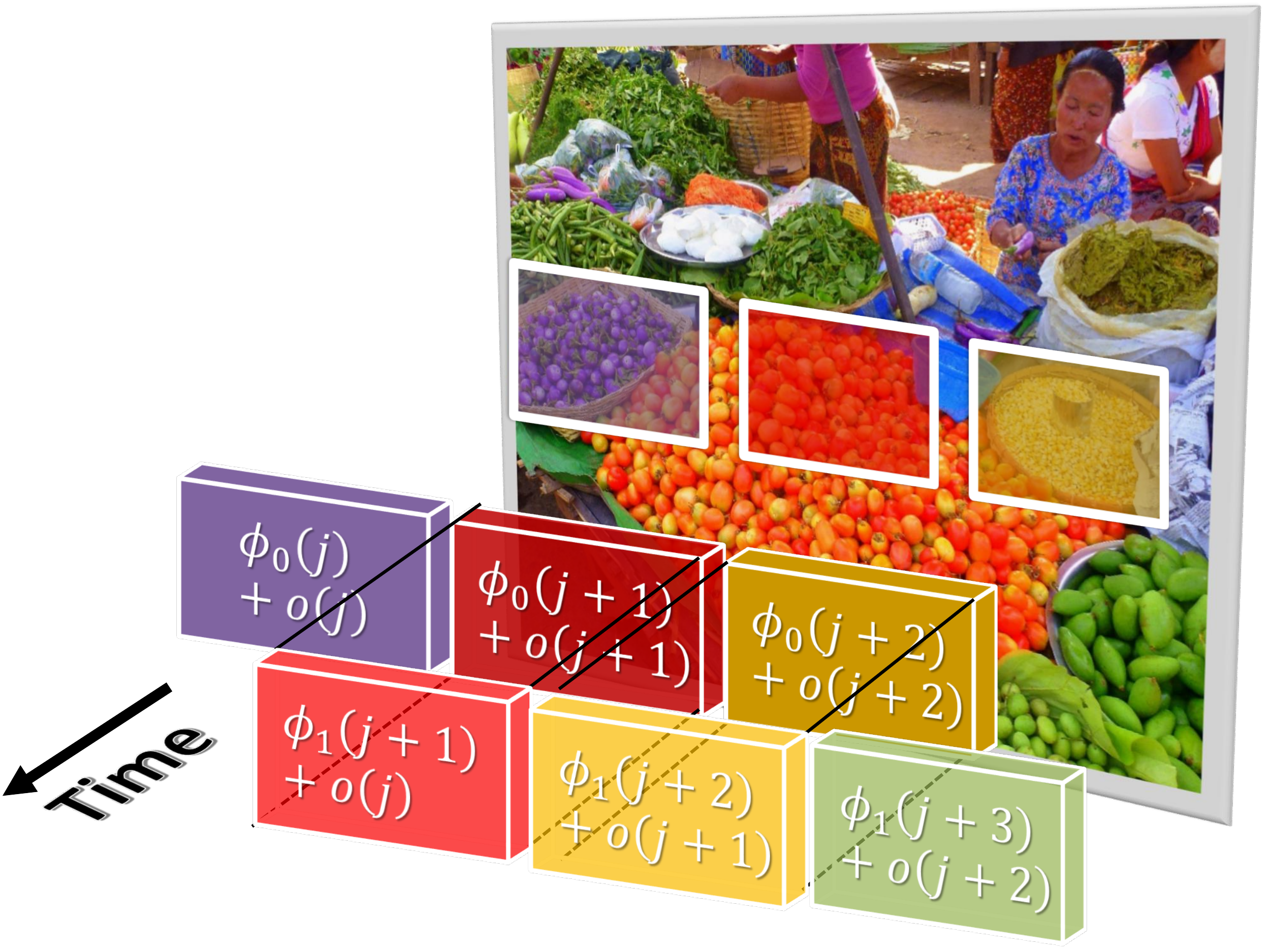}
\par\end{centering}
\caption{The scene as it captured before and after the transverse shift\label{fig:The-shifting-process}}
\end{figure}

After frame $n$ is captured, we physically shift the FPA by a single
pixel-sized step. Without lost of generality we shall first describe
dithering in the horizontal direction, leading to the next frame
\begin{equation}
R_{n+1}\left(i,j\right)=\phi_{n+1}\left(i,j+1\right)+o\left(i,j\right)\;.\label{eq:horizontal shift raw}
\end{equation}
The shifting process is described graphically in Fig. \ref{fig:The-shifting-process}.
 Based on frames $n$ and $n+1$, we can now calculate two differences:
\begin{itemize}
\item Temporal difference ($\Delta_{t}$) between raw frame $n+1$ (Eq.
\ref{eq:horizontal shift raw}) and raw frame $n$ (Eq. \ref{eq:gain compensated model})
\begin{eqnarray}
\Delta_{t}\left(R_{n}\left(i,j\right)\right) & \equiv & R_{n+1}\left(i,j\right)-R_{n}\left(i,j\right)\nonumber \\
\, & = & \phi_{n+1}\left(i,j+1\right)-\phi_{n}\left(i,j\right)\;.\label{eq:dx1}
\end{eqnarray}
\item Discrete horizontal derivative ($\Delta_{x}$) of raw frame $n$ (Eq.
\ref{eq:gain compensated model})
\begin{eqnarray}
\, & \, & \Delta_{x}\left(R_{n}\left(i,j\right)\right)\equiv R_{n}\left(i,j+1\right)-R_{n}\left(i,j\right)\nonumber \\
\, & \, & \qquad=\phi_{n}\left(i,j+1\right)-\phi_{n}\left(i,j\right)+o\left(i,j+1\right)-o\left(i,j\right)\;.\nonumber \\
\, & \, & \;\label{eq:dx2}
\end{eqnarray}
\end{itemize}
Subtraction of Eq. \ref{eq:dx1} from Eq. \ref{eq:dx2} results:

\begin{eqnarray}
\, & \, & \Delta_{x}\left(R_{n}\left(i,j\right)\right)-\Delta_{t}\left(R_{n}\left(i,j\right)\right)\nonumber \\
\, & \, & \qquad=\left[o\left(i,j+1\right)-o\left(i,j\right)\right]\nonumber \\
\, & \, & \qquad-\left[\phi_{n+1}\left(i,j+1\right)-\phi_{n}\left(i,j+1\right)\right]\;,
\end{eqnarray}
which can be rewritten as:

\begin{eqnarray}
\, & \, & \Delta_{x}\left(R_{n}\left(i,j\right)\right)-\Delta_{t}\left(R_{n}\left(i,j\right)\right)\nonumber \\
\, & \, & \qquad=\Delta_{x}\left(o\left(i,j\right)\right)-\Delta_{t}\left(\phi_{n}\left(i,j+1\right)\right)\;,\label{eq:dx3}
\end{eqnarray}
where $\Delta_{x}\left(o\left(i,j\right)\right)$ is the FPN discrete
horizontal derivative and $\Delta_{t}\left(\phi_{n}\left(i,j+1\right)\right)$
is the scene temporal difference. After frame $n+1$ is captured,
the FPA is shifted back to its original position.

Since $\Delta_{t}\left(\phi_{n}\left(i,j+1\right)\right)$ is usually
small for typical frame rate and since $\Delta_{x}\left(o\left(i,j\right)\right)$
and $\Delta_{t}\left(\phi_{n}\left(i,j+1\right)\right)$ are uncorrelated,
we can filter out $\Delta_{t}\left(\phi_{n}\left(i,j+1\right)\right)$
by computing Eq. \ref{eq:dx3} temporal median for several cycles,
leading to estimation of the FPN horizontal derivative $\Delta_{x}\left(o\left(i,j\right)\right)$. 

Similarly, the entire process is repeated for the vertical direction,
leading to an estimation of the FPN vertical derivative $\Delta_{y}\left(o\left(i,j\right)\right)$.

The last stage is the reconstruction of the FPN from the estimated
spatial derivatives. In one dimension, the reconstruction of a signal
from its derivatives is achieved by simple integration. The two dimensional
case is different since the estimated gradient vector field isn't
necessarily integrable. In other words, there might not exist a surface
such that $\left[\Delta_{x}\left(o\right),\Delta_{y}\left(o\right)\right]$
is its gradient field. In such a case, we seek for a surface $\tilde{o}\left(i,j\right)$
so that $\sum_{i,j}\left|\nabla\tilde{o}\left(i,j\right)-\left(\Delta_{x}\left(o\left(i,j\right)\right),\Delta_{y}\left(o\left(i,j\right)\right)\right)\right|$
is minimal. There are several algorithms addressing this problem,
including the projection of the estimated gradient field onto a finite
set of orthonormal basis functions and other iterative solvers \cite{key-100,key-101,key-102}.

Applying additional derivative on the gradient map and summing both
vector components, is equivalent to applying the Laplace operator
on the (\textit{a priory} unknown) offset map. In order to reconstruct
the FPN, we need to solve the Poisson equation:
\begin{equation}
\left(\Delta_{x},\Delta_{y}\right)\cdot\left(\Delta_{x}\left(o\left(i,j\right)\right),\Delta_{y}\left(o\left(i,j\right)\right)\right)^{\top}=\nabla^{2}o\left(i,j\right)
\end{equation}
This can be transformed into the frequency domain, by projecting the
estimated derivatives on an integrable set of functions, and solved
by integrating the projected set instead \cite{key-4}. Selection
of the specific set of integrable functions should be done according
to the problem constraints. We chose to project on the complete set
of cosines, that is the solution of the Poisson equation under Neumann
boundary conditions.

\begin{figure}[!tb]
\begin{centering}
\subfloat[Original frame\label{fig:Original-frame}]{\begin{centering}
\includegraphics[width=0.33\columnwidth]{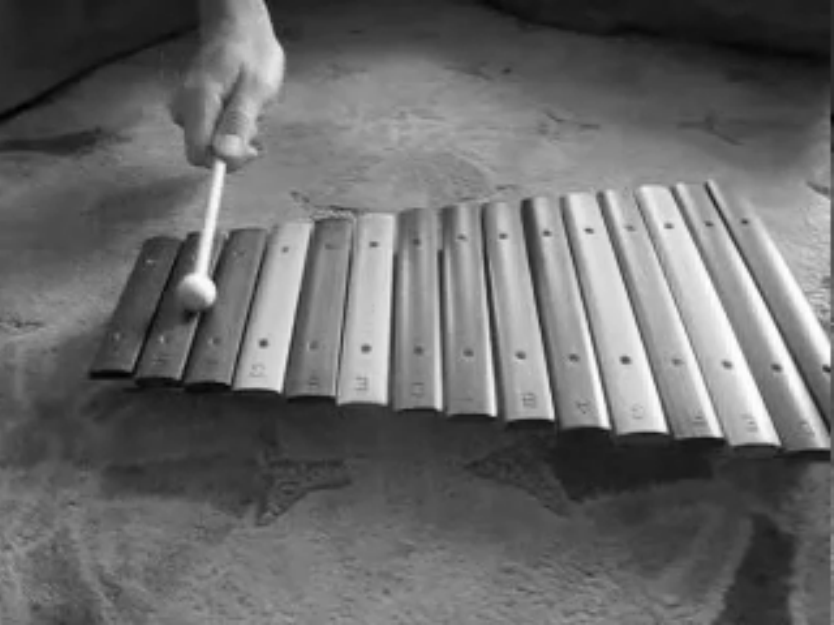}
\par\end{centering}
}\subfloat[Corrupted frame\label{fig:Corrupted-frame}]{\begin{centering}
\includegraphics[width=0.33\columnwidth]{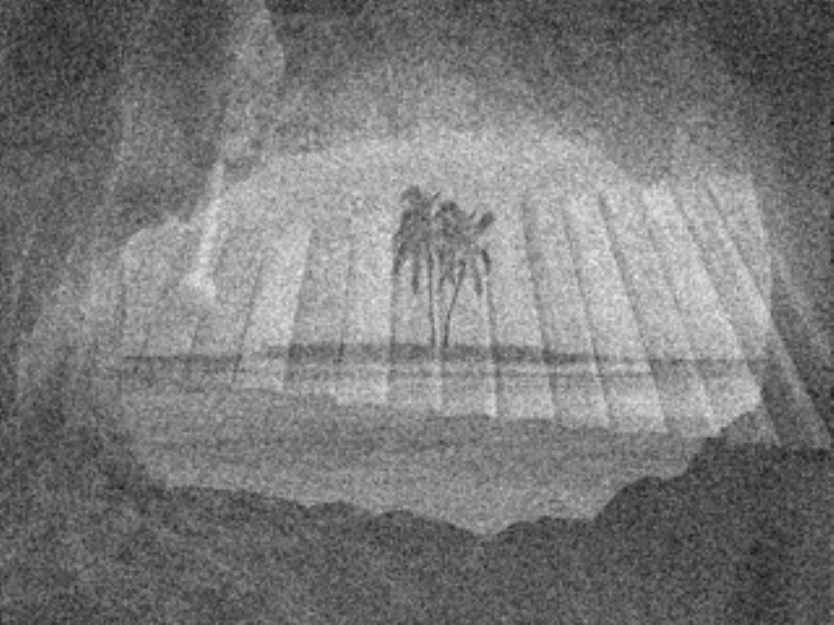}
\par\end{centering}
}\subfloat[Corrected frame\label{fig:Corrected-frame}]{\begin{centering}
\includegraphics[width=0.33\columnwidth]{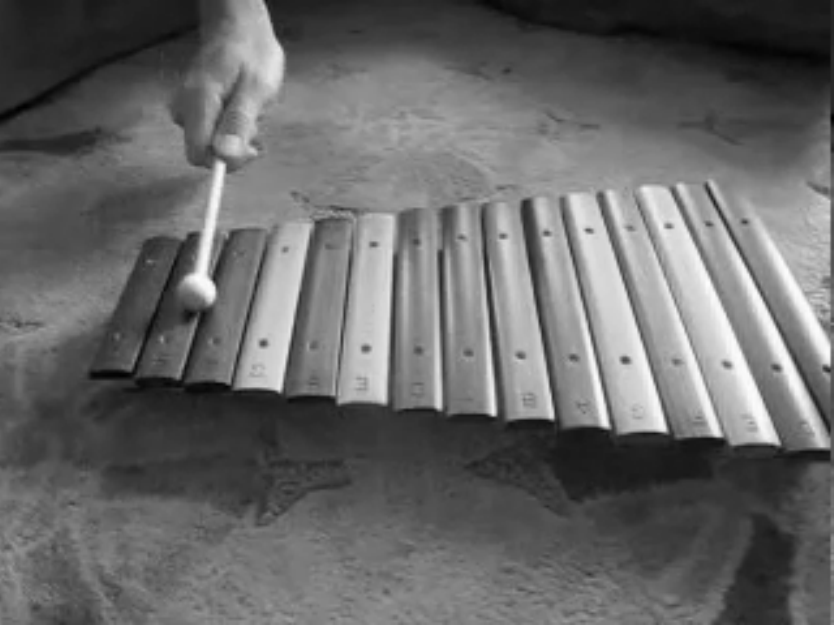}
\par\end{centering}
}
\par\end{centering}
\caption{Algorithm simulation\label{fig:Algorithm-simulation}}
\end{figure}

We have tested and measured the robustness of the method using simulated
dithering of virtual FPA. We used standard $240\times320$ pixels
video of normalized standard deviation. FPN was introduced by means
of sample picture which includes high and low spatial frequencies
as well as an extra constant random noise per pixel. The original
video frames were shifted over FPN as temporal random noise was added.
In Fig. \ref{fig:Algorithm-simulation}, we present original (Fig.
\ref{fig:Original-frame}), corrupted (Fig. \ref{fig:Corrupted-frame})
and corrected (Fig. \ref{fig:Corrected-frame}) arbitrary video frame.

\begin{figure}[!tb]
\begin{centering}
\subfloat[\label{fig:MSE-vs.-SNR,}]{\begin{centering}
\includegraphics[width=0.5\columnwidth]{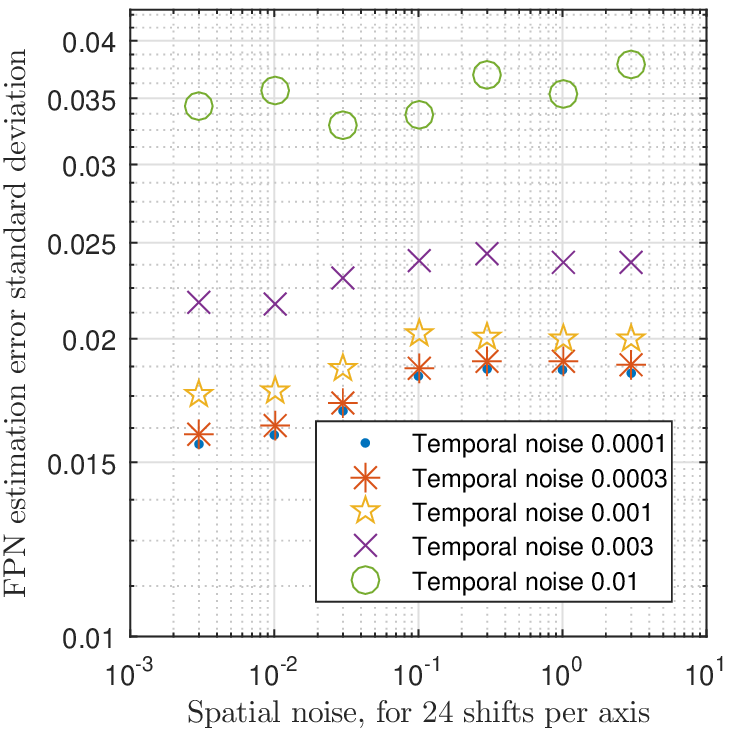}
\par\end{centering}
} \subfloat[\label{fig:MSE-vs.-number}]{\begin{centering}
\includegraphics[width=0.5\columnwidth]{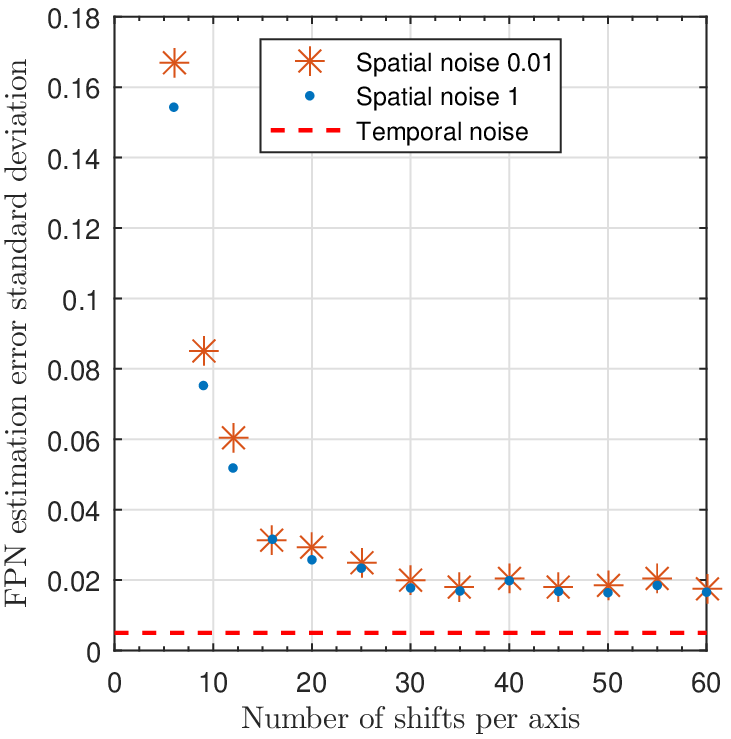}
\par\end{centering}
}
\par\end{centering}
\caption{\textbf{a}. Residual correction error vs. FPN strength, for several
temporal noises. One can notice the error kept small even for strong
FPN. \textbf{b}. Error vs. number of shifts, for several spatial noises.
One can notice the error decrease with the shifts number and the weak
dependence of spatial noise strength.}
\end{figure}

The residual error enumerated by standard deviation for the difference
between estimated and actual FPN, normalized by the standard deviation
of the scene. Error was calculated for various conditions of spatial
noise (FPN strength over original frame), temporal noise and number
of dithering repetitions, as in Fig. \ref{fig:MSE-vs.-SNR,} and \ref{fig:MSE-vs.-number}.
Notice that the error is only slightly dependent of the spatial noise
for various temporal noise (as expected since there was no assumption
for weak spatial noise), as can be noticed in Fig. \ref{fig:MSE-vs.-SNR,}
and \ref{fig:MSE-vs.-number}. The reconstruction performance increases
as more frames are used (Fig. \ref{fig:MSE-vs.-number}).

\begin{figure}[!tb]
\begin{centering}
\includegraphics[width=0.5\columnwidth]{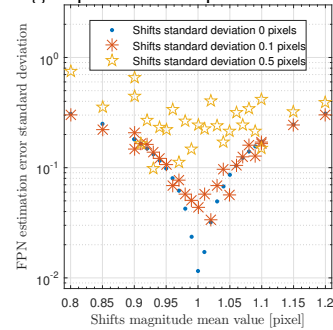}
\par\end{centering}
\caption{Residual error of the corrected FPN vs. applied shifts mean, for several
shifts standard deviations.\label{fig:MSE-by-mean}}
\end{figure}

Next we shall explore shift magnitude errors (as may caused by mechanical
apparatuses), in sub-pixel level. We used typical values of spatial
noise, temporal noise and number of shifts per axis ($0.1$, $0.0003$
and $32$ respectively), and compared the estimated FPN for different
mean and standard deviations of the translations. It should be mentioned
that the shift errors simulated as normally distributed random translations
(around the mean in the longitudinal direction and around zero in
the transverse direction). As shown in Fig. \ref{fig:MSE-by-mean},
the estimation seems to keep its accuracy even for significant shift
errors. 

We have demonstrated the method using microbolometer of $640\times480$
pixels, over complex and dynamic scene, which included both near and
far objects, having relative motion. High and low spatial frequencies
appeared in the scene. For convenience, the process previously introduced
was slightly modified, as the repetitive single pixel shift altered
by a constant angular velocity pitch and yaw of the entire imaging
system. This movement is equivalent to a transverse translations of
the FPA under the assumption of zero distortion system: If we set
the angular velocity to be the instantaneous field of view multiplied
by the frame rate, we get an effective single pixel shift per frame
of the imaged scene with respect to the sensor.

\begin{figure*}[t]
\begin{centering}
\subfloat[Raw signal\label{fig:Raw-signal}]{\begin{centering}
\includegraphics[width=0.33\textwidth]{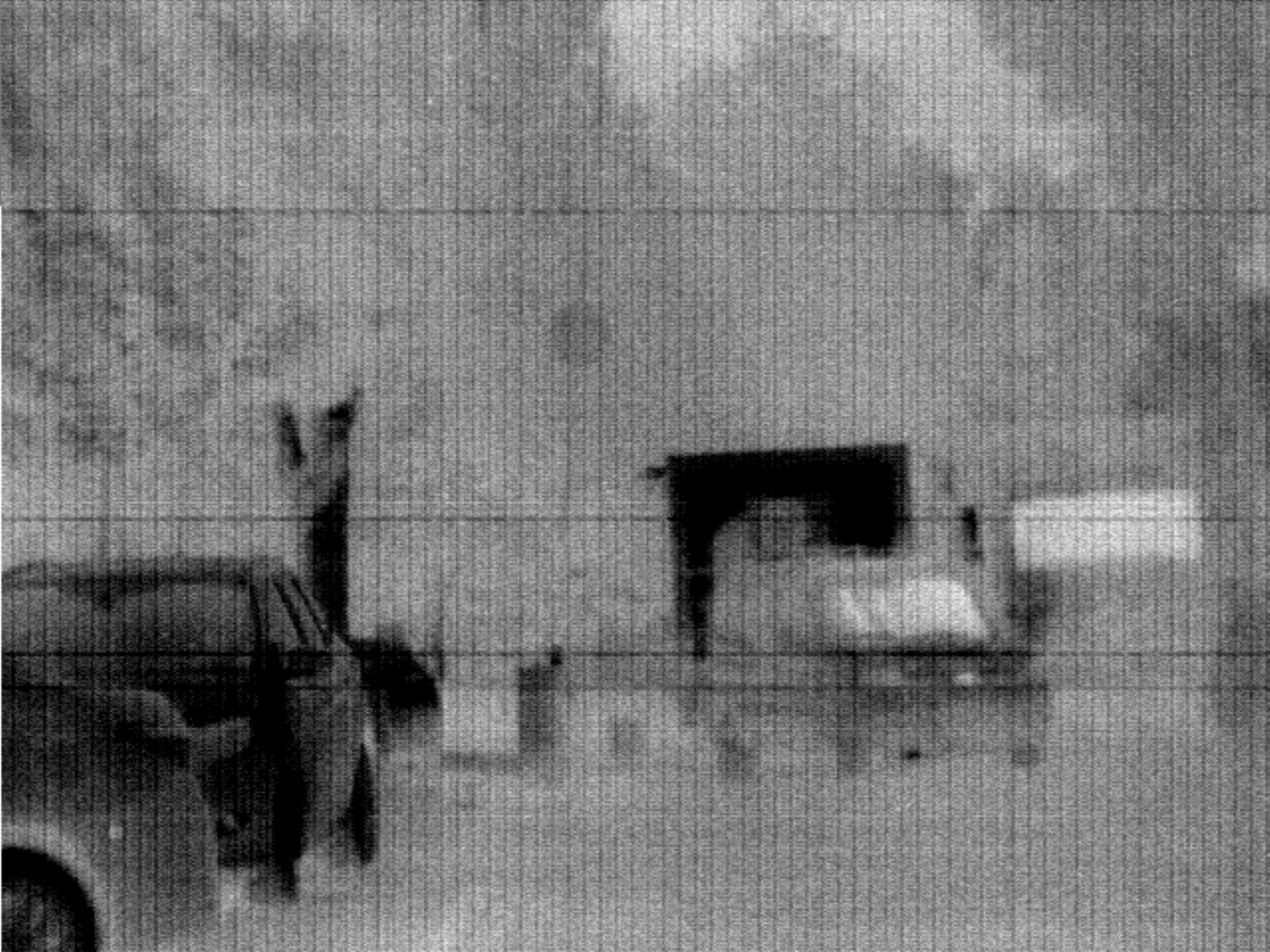}
\par\end{centering}
}\subfloat[Correction using proposed method\label{fig:Correction-using-our}]{\begin{centering}
\includegraphics[width=0.33\textwidth]{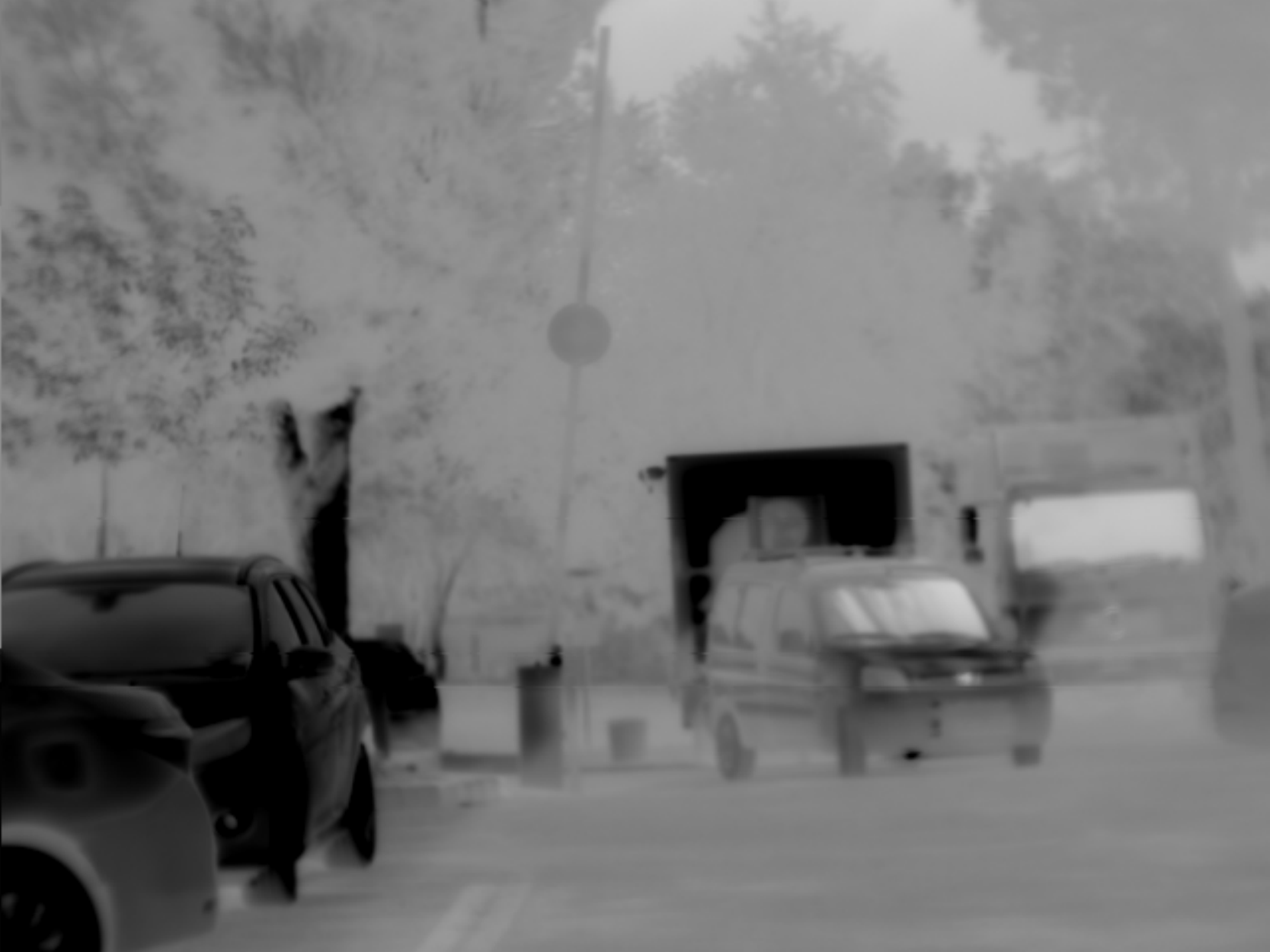}
\par\end{centering}
}\subfloat[Conventional correction\label{fig:Conventional-correction}]{\begin{centering}
\includegraphics[width=0.33\textwidth]{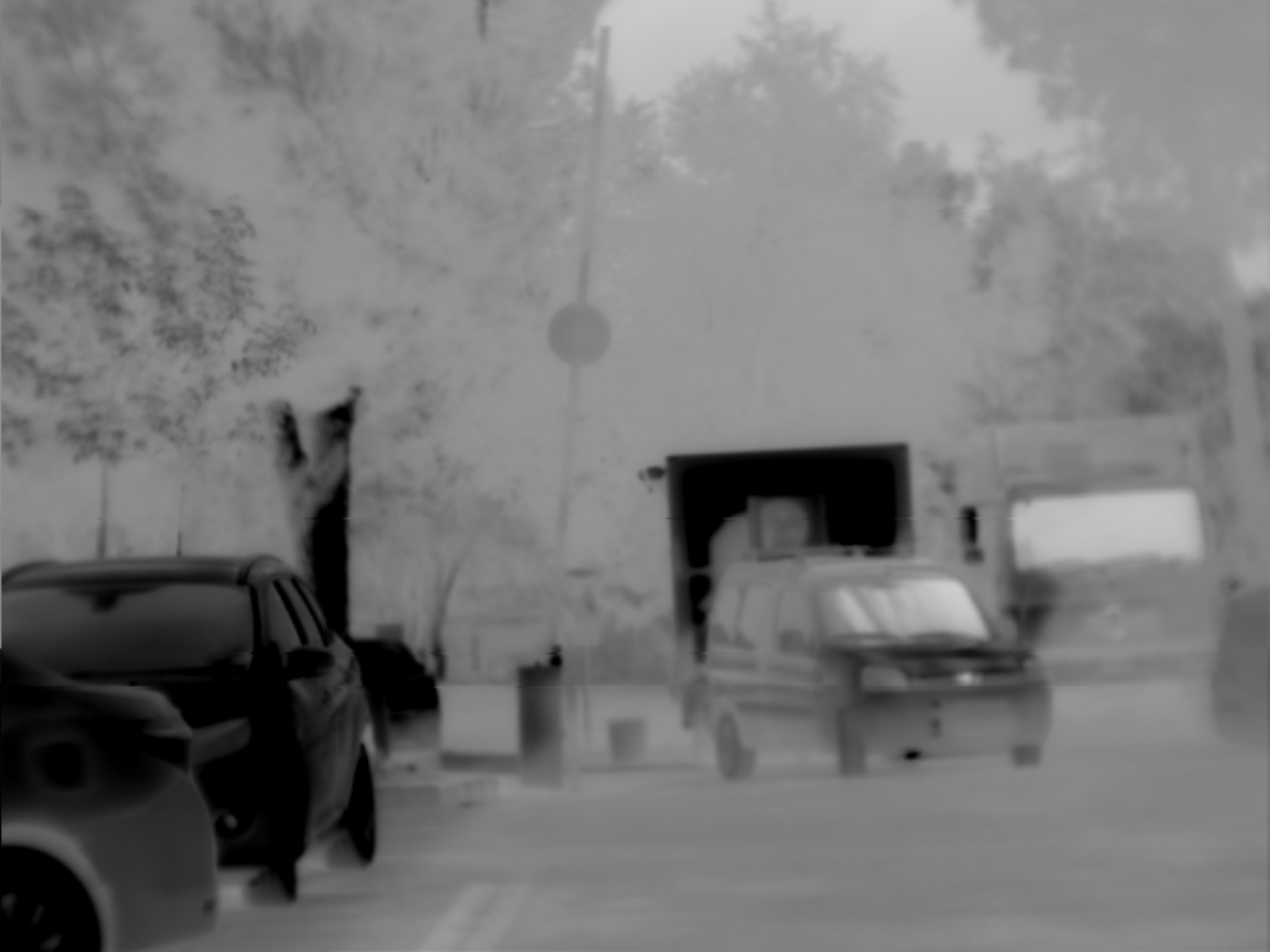}
\par\end{centering}
}
\par\end{centering}
\caption{Results comparison\label{fig:Results}}
\end{figure*}

In Fig. \ref{fig:Results}, we present the raw signal from the camera
(Fig. \ref{fig:Raw-signal}), the proposed method correction (Fig.
\ref{fig:Correction-using-our}) and the conventional bias and gain
correction (Fig. \ref{fig:Conventional-correction}). 

We present a method for FPN correction based on minute transverse
shifts of the FPA, which is robust to unexpected camera shakes and
interscene movements, provided with a demonstration of the algorithm
performance and accuracy in a complex scene. Implementation of this
algorithm can be applied using a transverse shift of the FPA or angular
movement of the optical axis. Although the proposed method has been
tested on uncooled microbolometer, it can be applied on any detector
suffering from either FPN or slowly varying noise of any magnitude
within the dynamic range, as long as single pixel shift is possible.
Therefore, it eliminates the need of long and costly calibrations
and mechanical mechanisms.

\section*{Acknowledgment}

The authors thank Ami Yaacobi, Eli Minkov and Ephi Pinnsky for their
support and assistance with this model. Special thanks also go to
Yohai Barnea, Pavel Bavly, Daniel Dahan and Yotam Ater for their technical
support.

\end{document}